# Effects of Introducing Synaptic Scaling on Spiking Neural Network Learning

Shinnosuke Touda[1]
Graduate School of Information Science and Technology
Osaka institute of technology
Osaka, Japan

Hirotsugu Okuno[2]
Fuculty of Information Science and Technology
Osaka institute of technology
Osaka, Japan
hirotsugu.okuno@oit.ac.jp

*Abstract*— Spiking neural networks (SNNs) employing unsupervised learning methods inspired by neural plasticity are expected to be a new framework for artificial intelligence. In this study, we investigated the effect of multiple types of neural plasticity, such as spike-time-dependent plasticity (STDP) and synaptic scaling, on the learning in a winner-take-all (WTA) network composed of spiking neurons. We implemented a WTA network with multiple types of neural plasticity using Python. The MNIST and the Fashion-MNIST datasets were used for training and testing. We varied the number of neurons, the time constant of STDP, and the normalization method used in synaptic scaling to compare classification accuracy. The results demonstrated that synaptic scaling based on the L2 norm was the most effective in improving classification performance. By implementing L2-norm-based synaptic scaling and setting the number of neurons in both excitatory and inhibitory layers to 400, the network achieved classification accuracies of 88.84 % on the MNIST dataset and 68.01 % on the Fashion-MNIST dataset after one epoch of training.

## I. Introduction

Spiking neural networks (SNNs), which consist of neuron models that generate pulsed potential changes, i.e., spikes, like the neurons in the brain, are expected to be a new framework for artificial intelligence that can perform flexible recognition like the brain. Unlike the formal neurons, which make up many conventional neural networks (NNs), spiking neurons have complex temporal properties like those of the brain, and therefore, SNNs generally require different learning rules than NNs consisting of formal neurons.

Among the learning rules occurring in the brain, spike-timing dependent plasticity (STDP) [1][2] and homeostatic plasticity (HP) [3] have long been the subject of research. STDP is a learning rule that updates synaptic weights depending on the difference in spike occurrence times between directly connected neurons. For example, in excitatory synapses, the connection strength is enhanced when a postsynaptic cell fires a spike within a particular time window after the spike firing of its presynaptic cell. Since this change is essentially positive feedback, a regulatory mechanism is essential to stabilize the network, and HP plays this role. HP includes synaptic scaling [4], which adjusts the overall weights of input synapses in a single neuron, and mechanisms that regulate excitability depending on the activity of the neuron (see [5][6] for overviews). The synergistic effects of STDP and HP have also been investigated in SNNs whose structure is different from the one used in this study [7][8].

The above learning rules work locally within a single neuron or between directly connected neurons. This point is different from backpropagation (BP) [9], a global learning rule widely used in formal-neuron-based NNs. Since BP requires information propagation in the reverse direction to the entire network, computational load for learning is large, and it is not easy to coexist BP in a network operated as a classifier. In contrast, the brain-inspired rules described above operate exclusively at the location and timing of spike firing, which alleviates computational overhead and allows the system to continue learning while simultaneously performing classification. In addition, these brain-inspired rules operate in an unsupervised manner and do not require labeled data for training.

Winner-take-all (WTA) networks have been used for implementing the brain-inspired learning rules. The WTA network consists of an excitatory and an inhibitory layer, and when properly trained by the above rules, neurons in the excitatory layer fire selectively to a particular input pattern. Previous studies have demonstrated unsupervised learning on several image datasets using WTA networks equipped with some of the brain-inspired learning rules ([10][11] for examples). Furthermore, efforts have been made to reduce the memory and energy requirements for implementing WTA networks in hardware [11]. Previous studies have also applied conditional processing lacking physiological plausibility to WTA networks in order to enhance learning efficiency and classification accuracy. These include reintroducing inputs after increasing the input spike frequency in response to insufficient spike output [10], and applying synaptic plasticity only to the neuron that exhibited the maximum firing for each training image [11].

In this study, we used the same network configuration as the previous study [10] and investigated the effects of introducing multiple types of brain-inspired learning rules. The main contributions of this study are the following two points. First, we demonstrated that the WTA network can be properly trained without involving physiologically unfeasible methods, such as re-entry with parameter changes. The elimination of such conditional processing also leads to faster training. Second, we demonstrated the effect of the function used for synaptic scaling on the classification accuracy. As mentioned above, STDP, which is essentially positive feedback, requires mechanisms that should work together to effectively regulate the increase in weights. This paper proposes appropriate complementary mechanisms to be used with STDP.

## II. RELATED WORKS

Methods for constructing SNNs that are applicable to classifiers can be broadly classified into the following two categories.

The first approach leverages findings from DNN research based on formal neuron models. These methods include converting formal neurons in trained DNNs into spiking neurons [12][13], as well as adapting BP [9] for training SNNs [14][15][16]. Such methods typically exhibit a certain degree of performance degradation compared to the original DNNs prior to conversion.

The second method is to construct SNNs using brain-inspired learning rules, such as STDP and HP. Currently, NNs developed using this method show inferior accuracy to NNs using DNN-based methods, but this method can take advantage of the benefits of brain-inspired learning rules described in section I. Diehl et al. demonstrated that a WTA network based on brain-inspired learning rules can classify handwritten digits with sufficient accuracy [10]. SNNs that achieve higher accuracy by including convolutional layers and by including adaptive synapse computation, threshold, and lateral inhibition have also been reported [17][18].

This study follows the latter method. The scope of this study is limited to the classification component of a NN architecture, and therefore, the network used in this study does not have any feature extraction components, such as convolutional neural networks.

## III. NEURAL NETWORK AND NEURON-SYNAPSE MODELS

### A. Neural Network Architecture

Fig. 1 shows the WTA network used in this study, which consists of an input layer, an excitatory layer, and an inhibitory layer. The input layer consists of 784 neurons, each of which corresponds to a pixel in the input image. Neurons in the input layer are fully coupled to neurons in the excitatory layer. The excitatory and inhibitory layers have the same number of neurons, 100 or 400. Each neuron in the excitatory layer is connected to a neuron in the inhibitory layer and inhibits neurons in the excitatory layer other than itself through neurons in the inhibitory layer.

In the WTA network, winners are selected by the following manner. Each neuron in the excitatory layer inhibits each other via inhibitory neurons. This results in the suppression of spike firing except for the "winner neurons", which present a spike rate higher than a certain rate. In this situation, STDP, which is explained in section IV, increases the input weights for the winner neurons, and the difference in the number of spikes between the winner neurons and the other neurons gradually increases.

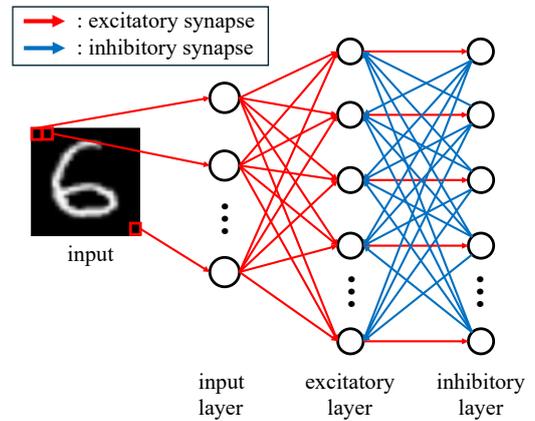

Fig. 1. Winner-take-all network architecture.

### B. Neuron Models

This section describes the two types of neuron models used in the NN designed in this study. For the input layer neurons, a model in which the number of spikes follows a Poisson distribution was employed. When the probability of occurrence of an event follows a Poisson distribution with expected value $\lambda$, the probability of this event occurring once during a short time interval $\Delta t$ is expressed by the following equation:

$$P[X(\Delta t) = 1] = \frac{(\lambda \Delta t)^1 e^{-\lambda \Delta t}}{1!} \simeq \lambda \Delta t. \tag{1}$$

In the implemented program, a random number following a uniform distribution $U(0,1)$ is generated and a spike is generated if its value is less than $\lambda \Delta t$.

For the excitatory and inhibitory layers, the leaky integrate-and-fire (LIF) model was employed as the neuron model. Fig. 2(a) shows the circuit model of the LIF neuron. $v_i$ represents the membrane potential of neuron $i$. $C_m$ and $g_m$ represent the capacitance and the conductance of the cell membrane, respectively. $I_S$ represents the synaptic current of chemical synapses, and $E_L$ represents the resting membrane potential. The membrane potential varies according to the following differential equation:

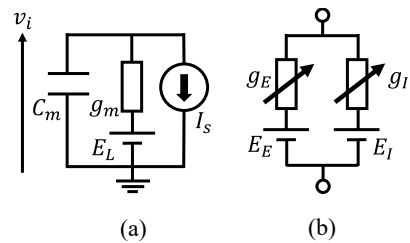

Fig. 2. Circuit models of the neuron and the synapse. (a) Circuit model of neurons in the excitatory and inhibitory layers. (b) Circuit model of the chemical synapse.

$$C_m \frac{dv_i}{dt} = -g_m(v_i - E_L) + I_S. \quad (2)$$

When the membrane potential reaches a certain threshold $v_\theta$, the neuron generates a spike. The membrane potential decreases to the reset potential at the next step of firing and enters a period during which the membrane potential does not change, i.e., the refractory period; the length of the period is $t_r$. The thresholds of excitatory layer neurons change with learning. The details are described in section IV. The thresholds of inhibitory neurons are constant $v_\theta = v_{\theta 0}$.

Table I shows the parameters of neurons in the excitatory and inhibitory layers. $\tau_m$ in the table represents the time constant of neurons.

*C. Synapse Model*

Fig. 2(b) shows the circuit model of the chemical synapse $I_S$. $g_{E,k}$ and $g_{I,k}$ represent excitatory and inhibitory synaptic conductance whose presynaptic neuron is $k$, respectively, and $E_E$ and $E_I$ represent excitatory and inhibitory equilibrium potentials, respectively. The total current $I_S$ of the excitatory and inhibitory chemical synapses are represented by

$$I_S = \sum_{k=1}^{N_E} g_{E,k}(v_i - E_E) + \sum_{k=1}^{N_I} g_{I,k}(v_i - E_I), \quad (3)$$

where $N_E$ and $N_I$ denote the total number of excitatory and inhibitory presynaptic neurons, respectively. The model of synaptic current used here is the single exponential synapse model, in which the synaptic conductance ($g_{E,k}$ and $g_{I,k}$) of the postsynaptic neuron increases at the time of spike firing of the presynaptic neuron and then decreases exponentially. Let $t_{pre}$ be the time of spike firing of the presynaptic neuron, the excitatory synaptic conductance $g_{E,k}$ of the postsynaptic neuron is expressed by

$$g_{E,k}(t) = \frac{w_{E,k}}{\tau_s} \exp\left(-\frac{t - t_{pre}}{\tau_s}\right), \quad (4)$$

where $w_{E,k}$ is the excitatory synaptic weight with presynaptic neuron $k$. The inhibitory conductance $g_{I,k}$ is calculated in the same way.

Table II shows parameters of the synapses connected to neurons in each layer.

## IV. LEARNING RULES

*A. Spike-Timing Dependent Plasticity*

Synaptic weights from the input layer to the excitatory layer are updated by a combination of STDP and HP, which includes synaptic scaling and excitability modulation.

TABLE I. PARAMETERS OF EXCITATORY AND INHIBITORY LAYER NEURONS

|  | Excitatory Layer | Inhibitory Layer |
|---|---|---|
| $v_{\theta 0}$ | -40 mV | -40 mV |
| $E_L$ | -60 mV | -60 mV |
| $t_r$ | 2 ms | 5 ms |
| $\tau_m (= C_m/g_m)$ | 30 ms | 5 ms |
| $\tau_\theta$ | $2 \times 10^5$ ms | - |
| $\alpha$ | 0.11 mV | - |

TABLE II. PARAMETERS OF SYNAPSES CONNECTED TO NEURONS IN EACH LAYER

|  | Excitatory Layer | Inhibitory Layer |
|---|---|---|
| $E_E$ | 0 mV | 0 mV |
| $E_I$ | -80 mV | - |
| $\tau_s$ | 20 ms | 20 ms |

STDP is a learning rule found in the biological neuronal network, and updates weights depending on the difference in spike firing time $\Delta t_f (= t_{post} - t_{pre})$ between presynaptic and postsynaptic neurons [1][2]. The change in weights $\Delta w$ in this learning rule is expressed by the following equation:

$$\Delta w = \begin{cases} A_+ \exp\left(\frac{\Delta t_f}{\tau_p}\right) & \text{if } \Delta t > 0 \\ -A_- \exp\left(-\frac{\Delta t_f}{\tau_p}\right) & \text{if } \Delta t < 0 \end{cases}. \quad (5)$$

Here, $A_+$ and $A_-$ correspond to the learning rate. The values used in this study are the followings: $A_+ = 0.005$, $A_- = 0.0025$. Fig. 3 shows the relationship between $\Delta t_f$ and $\Delta w$.

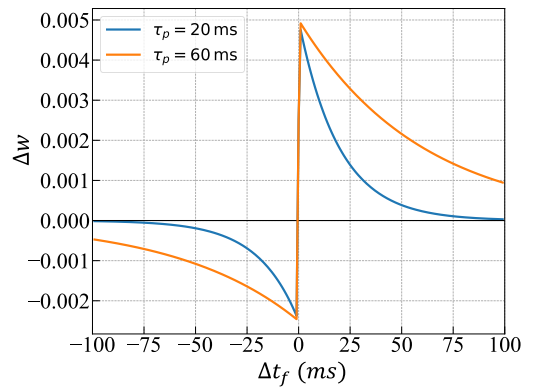

Fig. 3. Relationship between the difference in spike firing timing $\Delta t_f$ and change in weight $\Delta w$.

## B. Synaptic Scaling

In addition to STDP, synaptic scaling also updates the synaptic weights. While STDP updates individual synaptic weights, synaptic scaling updates all synaptic weights input to a single neuron. The synaptic scaling implemented in this study normalizes the weight vector $w$ that represents weights from the input layer to a neuron in the excitatory layer according to the following formula:

$$w' = \frac{W_t}{\|w\|_p} w, \qquad (6)$$

where $\|w\|_p$ represents the $L_p$ norm, and the L1 and L2 norms were used. The coefficient $W_t$ represents the target norm and corresponds to the norm in a steady state where the weights remain unchanged.

## C. Excitability Regulation

Since STDP-based weight updating is essentially positive feedback, without proper control, only the weights of certain neurons will increase, and firing will be monopolized. To prevent monopoly, neuronal excitability regulation is required along with STDP. While there are various ways to regulate excitability, excitability was regulated by changing the firing threshold for simplicity in this study.

The firing threshold of a neuron in the excitatory layer is defined as $v_\theta = v_{\theta 0} + \theta$, where $v_{\theta 0}$ is a constant representing the initial threshold and $\theta$ changes dynamically according to the following differential equation:

$$\frac{d\theta}{dt} = -\frac{\theta}{\tau_\theta} + \alpha \delta(t - t_f), \qquad (7)$$

where $\delta$ is the Dirac delta function and $t_f$ is the firing time of the neuron. The firing threshold of a neuron in the excitatory layer increases $\alpha$ mV with each firing and decays with a time constant $\tau_\theta$.

## V. SIMULATION METHOD

### A. Simulation Settings

We implemented the above WTA network and learning rules using Python 3.12. The time step of the simulation ($\Delta t$) was set to 1 ms. The datasets used for training and testing were MNIST [19], which is an image dataset consisting of handwritten digits from 0 to 9, and Fashion-MNIST [20], which is also an image dataset consisting of 10 classes of clothing and accessories.

Each image in both datasets consists of 784 pixels (28 × 28). We ran three simulations with different initial weights, and the average value of the results of three simulations is the subject of our evaluation.

For training the SNN, 40,000 images, 4,000 for each class, were randomly selected from 60,000 training images in MNIST or Fashion-MNIST. The presentation time per input image was 150 ms. The initial values of synaptic weights from the input layer to the excitatory layer were random numbers following a uniform distribution $U(0,1)$. For testing, 10,000 test images were used. The presentation time per input image was 500 ms. During training and testing, each neuron's membrane potential was reset to its resting membrane potential each time the input image was switched.

### B. Classification Accuracy

To evaluate classification accuracy, each spike output patterns should be associated with the class predicted by the network. Class labels of the training dataset were used solely for this association process. First, the average firing rate $m_{il}$ of each neuron $i$ in the excitatory layer for each class $l \in \{0, 1, \cdots, 9\}$ is calculated for all training images. The class $C_i$ assigned to neuron $i$ is determined according to the following rule:

$$C_i = \underset{l}{\mathrm{argmax}}\, m_{il}. \qquad (8)$$

Next, the predicted class for an input image $j$ is determined by the label assigned to the neuron that exhibits the highest spike count in response to image $j$. This estimation method, which relies solely on the neuron exhibiting the highest firing rate, demonstrated higher accuracy than the approach used in previous study [10], where the predicted class is determined based on the average firing rate of neurons assigned to each class, particularly when the network contains a large number of neurons.

## VI. RESULTS

### A. Effects of the Method for Synaptic Scaling

The classification accuracy of methods employing L1 and L2 norms in synaptic scaling was investigated on the MNIST and Fashion-MNIST datasets. The number of neurons in each of the excitatory and inhibitory layers was set to either 100 or 400, and the STDP time constant $\tau_p$ was set to either 20 ms or 60 ms. The classification accuracy was compared across different values of $W_t$. Tables III and IV respectively show the classification accuracy of methods using the L1 norm and L2 norm on the MNIST dataset, while Tables V and VI respectively show the classification accuracy of methods using the L1 norm and L2 norm on the Fashion-MNIST dataset. The red numbers in the table indicate the highest value in each row.

These results indicate that the classification accuracy of methods using the L2 norm is consistently higher than those using the L1 norm regardless of the dataset used. The underlying reason for this observation may lie in the post-training weights shown in Fig. 4 in the next section.

The classification accuracy was found to be highly dependent on the parameter $W_t$, and the $W_t$ value that yielded the highest classification accuracy varied depending on the number of neurons and the value of $\tau_p$. Since $W_t$ represents the L1 or L2 norm of the weight vector at the stable state after synaptic weight updates, this means that the optimal value of the norm in the stable state depends on various parameters. Furthermore, when $W_t$ was set too small, the lack of sufficient weight prevented learning from progressing, and class-specific weight patterns, as shown in the next section, were not formed. The very low classification accuracy at $W_t = 60$ in Table V is the result of this. These facts suggest the necessity of methods that allow $W_t$ to also change through learning.

TABLE III. CLASSIFICATION ACCURACY (%) OF THE MNIST DATASET WITH L1 NORM SYNAPTIC SCALING.

| Neuron Number | $\tau_p$ | $W_t$ | | | | |
|---|---|---|---|---|---|---|
| | | 60 | 80 | 120 | 160 | 200 |
| 100 | 20 | 53.17 | 72.70 | 77.02 | 78.57 | 68.94 |
| | 60 | 61.08 | 73.57 | 81.18 | 83.61 | 81.73 |
| 400 | 20 | 57.90 | 74.02 | 79.10 | 78.51 | 69.68 |
| | 60 | 63.38 | 74.88 | 85.63 | 87.05 | 85.02 |

TABLE IV. CLASSIFICATION ACCURACY (%) OF THE MNIST DATASET WITH L2 NORM SYNAPTIC SCALING.

| Neuron Number | $\tau_p$ | $W_t$ | | | | |
|---|---|---|---|---|---|---|
| | | 3 | 5 | 7 | 9 | 11 |
| 100 | 20 | 83.46 | 82.74 | 81.76 | 80.78 | 73.41 |
| | 60 | 85.21 | 85.08 | 85.31 | 86.30 | 84.66 |
| 400 | 20 | 84.25 | 84.01 | 82.62 | 81.90 | 76.47 |
| | 60 | 85.80 | 88.06 | 88.58 | 88.84 | 86.96 |

TABLE V. CLASSIFICATION ACCURACY (%) OF THE FASHION-MNIST DATASET WITH L1 NORM SYNAPTIC SCALING.

| Neuron Number | $\tau_p$ | $W_t$ | | | | |
|---|---|---|---|---|---|---|
| | | 60 | 80 | 120 | 160 | 200 |
| 100 | 20 | 12.50 | 50.40 | 45.42 | 44.69 | 41.20 |
| | 60 | 17.83 | 50.50 | 45.73 | 43.62 | 40.05 |
| 400 | 20 | 19.64 | 50.21 | 43.79 | 40.55 | 38.59 |
| | 60 | 22.42 | 48.47 | 43.14 | 42.85 | 38.74 |

TABLE VI. CLASSIFICATION ACCURACY (%) OF THE FASHION-MNIST DATASET WITH L2 NORM SYNAPTIC SCALING.

| Neuron Number | $\tau_p$ | $W_t$ | | | | |
|---|---|---|---|---|---|---|
| | | 3 | 5 | 7 | 9 | 11 |
| 100 | 20 | 64.14 | 52.86 | 54.40 | 48.91 | 47.90 |
| | 60 | 51.92 | 56.67 | 53.89 | 53.28 | 54.68 |
| 400 | 20 | 68.01 | 57.32 | 51.05 | 48.55 | 48.04 |
| | 60 | 65.02 | 54.67 | 52.45 | 50.93 | 50.18 |

Regarding the number of neurons, when $W_t$ was appropriately set in the L2-norm-based method, a network comprising 400 neurons consistently outperformed that comprising 100 neurons across datasets. This positive correlation between the number of neurons and classification accuracy aligns with findings reported in previous studies.

On the other hand, the value of $\tau_p$ that yielded superior classification accuracy varied depending on the dataset. For the MNIST dataset, setting $\tau_p$ to 60 ms resulted in higher accuracy, whereas for the Fashion-MNIST dataset, $\tau_p$ of 20 ms led to better performance. Since the optimal $\tau_p$ is expected to depend on the statistical properties of the input spikes, further experiments from this perspective are warranted.

The method using the L2 norm achieved the highest classification accuracy of 88.84 % on the MNIST dataset under the following conditions: the number of neurons = 400, $\tau_p$ = 60 ms, and $W_t$ = 9. Similarly, the L2 norm method showed the highest classification accuracy of 68.01 % on the Fashion-MNIST dataset under the following conditions: the number of neurons = 400, $\tau_p$ = 20 ms, and $W_t$ = 3. This value for the Fashion-MNIST surpasses the accuracy of the previous studiy that used unsupervised learning with STDP [9].

### B. Self-organization of Synaptic Weight during Training

Fig. 4 shows how the synaptic weights are organized during training. Figs. 4(a) and 4(b) show the weights obtained using methods that employ the L1 and L2 norm, respectively, on the MNIST dataset. Fig. 4(c) shows the weights obtained using a method employing the L2 norm with a low $W_t$ value ($W_t$ = 2), which inhibits learning progress, also on the MNIST dataset. Fig. 4(d) shows the weights obtained using a method employing the L2 norm on the Fashion-MNIST dataset. The weights in Figs. 4(a), (b), and (d) were obtained using the parameters that yielded the highest classification accuracy. As shown in these figures, the weights start from random numbers and gradually approaches to the form of one of the input pattern.

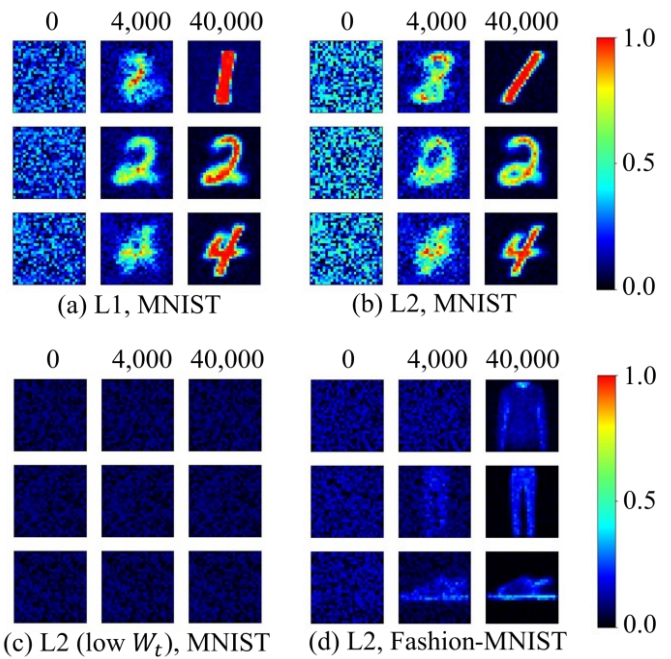

Fig. 4. Time lapse images showing the synaptic weights during training. The numbers at the top of the figure represent the number of input images used for training. (a) and (b) respectively represent the weights obtained using methods based on the L1 norm and the L2 norm on the MNIST dataset. (c) Weights obtained using a method employing the L2 norm with a low $W_t$ value ($W_t$ = 2), which inhibits learning process, on the MNIST dataset. (d) Weights obtained using a method employing the L2 norm on the Fashion-MNIST dataset.

Comparing Figs. 4(a) and 4(b), the weights obtained using the L1-norm-based method are generally higher than those obtained using the L2-norm-based method. This relationship is also evident in the statistics of all neuron weights beyond the examples presented (not included in this paper). While this difference may affect classification accuracy, further investigation is required to determine the underlying cause.

## VII. Conclusion

In the present study, we investigated the impact of introducing synaptic scaling into training of a WTA network, as well as the effect of normarizing methods of synaptic scaling on classification accuracy.

The comparison of normalizing methods showed that the method using the L2 norm gives higher classification accuracy than that using the L1 norm. Without relying on physiologically unfeasible procedures, such as re-entry with parameter changes, our network, which employs an L2-norm-based approach, achieved a classification accuracy of 88.84 % on the MNIST dataset and 68.01 % on the Fashion-MNIST dataset. The formation of weights contributing to classification required a $W_t$ value above a certain threshold, whereas the STDP time constant had little influence on this process. Additional research is necessary to elucidate the reasons why the L2-norm-based method demonstrates superior performance compared to other approaches.

## Acknowledgment

This work was supported by JSPS KAKENHI Grant Number JP24H02338. We are grateful to Dr. Akito Morita for his valuable advice regarding the Python-based implementation of the spiking neural network.